\documentclass[runningheads]{llncs}
\usepackage{graphicx}
\usepackage{amssymb}
\usepackage{subcaption}
\usepackage{wrapfig}
\usepackage{booktabs}
\usepackage{amsmath}
\usepackage[misc]{ifsym}
\usepackage{hyperref,xcolor}

\usepackage{subfiles}
\usepackage{pifont}
\newcommand{\cmark}{\ding{51}}%
\newcommand{\xmark}{\ding{55}}%

\begin{document}
\title{S-SAM: SVD-based Fine-Tuning of Segment Anything Model for Medical Image Segmentation
}
%
%
%
%

%
\author{Jay N. Paranjape\inst{1,}\Letter \and
Shameema Sikder\inst{2,3} \and
S. Swaroop Vedula\inst{3} \and
Vishal M. Patel\inst{1}
}
\authorrunning{J. Paranjape et al.}
\institute{Department of Electrical and Computer Engineering, The Johns Hopkins University, Baltimore, USA \\
\email{jparanj1@jhu.edu}
\and
Wilmer Eye Institute, The Johns Hopkins University, Baltimore, USA \and Malone Center for Engineering in Healthcare, The Johns Hopkins University, Baltimore, USA
}
\maketitle              
%
\begin{abstract}
Medical image segmentation has been traditionally approached by training or fine-tuning the entire model to cater to any new modality or dataset. However, this approach often requires tuning a large number of parameters during training. With the introduction of the Segment Anything Model (SAM) for prompted segmentation of natural images, many efforts have been made towards adapting it efficiently for medical imaging, thus reducing the training time and resources. However, these methods still require expert annotations for every image in the form of point prompts or bounding box prompts during training and inference, making it tedious to employ them in practice. In this paper, we propose an adaptation technique, called S-SAM, that only trains parameters equal to 0.4\% of SAM's parameters and at the same time uses simply the label names as prompts for producing precise masks. This not only makes tuning SAM more efficient than the existing adaptation methods but also removes the burden of providing expert prompts. We call this modified version S-SAM and evaluate it on five different modalities including endoscopic images, x-ray, ultrasound, CT, and histology images. Our experiments show that S-SAM outperforms state-of-the-art methods as well as existing SAM adaptation methods while tuning a significantly less number of parameters. We release the code for S-SAM at \url{https://github.com/JayParanjape/SVDSAM}.


\end{abstract}

\section{Introduction}
Image segmentation is a fundamental task in medical image analysis. Various deep learning (DL) based algorithms have been proposed in the literature that are able to perform remarkably well on a wide variety of modalities including CT, MRI, X-ray, ultrasound and endoscopy for segmenting organs, tumors, and tissues \cite{dl3,dl5_survey}. However, most of the well-performing models inherently have a large number of parameters, thus increasing the resources needed to train them every time for a new dataset or a new modality. This problem has been tackled for natural images to some extent by foundational models which are trained on billions of data points, exhibit an innate understanding of a given task, and are able to generalize well, given suitable user prompts. Notable examples include CLIP \cite{clip} and ALIGN \cite{align} for image classification and open set image-text understanding, and the recently proposed Segment Anything Model (SAM) \cite{sam} for segmentation. As such, it was hoped that medical image segmentation could also be addressed using SAM. However, SAM is not trained on medical images and various studies show its inability to generalize well on medical data \cite{failsam2,adaptivesam}. 
In order to utilize the power of SAM on medical images, there have been various efforts to adapt it efficiently \cite{medsam,medsamadapter,sonosam,samus}. However, it is not feasible to provide expert-level prompts for every image during training or inference, as is required by these adaptation methods since this process is time-consuming for experts. In our work, we propose S-SAM - an adaptation of SAM that only expects the name of the class of interest as a prompt, thus eliminating the need for expert-level prompts. This property makes it more suitable for usage in medical systems while also retaining the promptable nature of SAM. Furthermore, we show that our approach is significantly more efficient and requires training far fewer parameters than existing methods. In the proposed approach, we tune the singular values of the weight matrices in SAM's image encoder. Singular values define how important each visual feature in the image is to the activation of a given layer of the model \cite{eigenfaces} and hence, modifying these during training allows S-SAM to correctly adapt SAM for a given task.

In summary, the main contributions of our paper are as follows. (1) We propose a novel adaptation of SAM, called S-SAM, that can perform text-prompted segmentation of medical images. S-SAM expects prompts as simple as the name of the class of interest to produce precise masks, thus removing the requirement of expert-level prompts. (2) We develop a technique for training S-SAM that makes it far more efficient than existing adaptation methods. In comparison with the original SAM, the number of trainable parameters for S-SAM are only 0.4\%. (3)  Extensive experiments are conducted on five publicly available datasets of different modalities where we obtain the SOTA performance.
\section{Related Work}
While SAM is a powerful tool for natural image segmentation, it needs to be adapted to perform well in the medical domain. Hence, much research has been conducted to harness the power of SAM's encoders and decoders for medical images. One of the first efforts in this direction was MedSAM \cite{medsam}, where the authors finetune SAM for a huge corpus of medical images and provide support for point and bounding box-based prompts. While this approach finetunes all the parameters of SAM, the Medical SAM Adapter \cite{medsamadapter} introduces learnable adapter layers for all the encoder and decoder blocks. These are low-rank approximations that allow learning the domain shift while freezing all other parameters of SAM. While this method significantly brings down the GPU requirement from 1024 (for SAM) to 4, it can still be considered compute-intensive. SonoSAM \cite{sonosam} tunes the mask decoder and prompt encoder of SAM for training on sonography images. SAMUS \cite{samus}, on the other hand, adapts SAM for ultrasound images by training an additional CNN-based image encoder and fusing it with SAM's encoder. We refer readers to a comprehensive survey of SAM-adaptation methods \cite{sam_survey} for more details. However, these approaches still require experts to manually provide precise point prompts or bounding box prompts for every image during training and testing, which is tedious. Hence, an adaptation of SAM that minimizes expert involvement would be highly beneficial for medical applications. In light of this, SAMed \cite{samed} was introduced, which added trainable low-rank adaptation layers to SAM's image encoder while keeping the rest of the encoder weights frozen. In addition, the user-defined prompts were replaced with a default prompt to eliminate expert involvement. Similarly, AutoSAM \cite{autosam} replaces the user-defined prompt with the image itself, and trains the prompt encoder to produce good prompt embeddings with the image input. However, these techniques remove the promptable nature of SAM. To tackle this, AdaptiveSAM \cite{adaptivesam} introduces text-prompted segmentation which allows the use of label names as the prompt. 
\begin{wraptable}{r}{0.65\textwidth}
\centering
\caption{Comparison of different adaptation methods of SAM present in the literature. Here, expert intervention means providing expert prompts for every image, including points, bounding boxes or masks. We exclude text prompts with label names from this category since experts are not required to provide the label name prompts for every image, but only once for the entire dataset.}
\setlength{\tabcolsep}{3pt}
\resizebox{0.65\textwidth}{!}{%
\begin{tabular}
{|c|c|c|c|}
\hline
Method & Expert Intervention Not Required & Promptable & Training Decoder Not Required\\
\hline
MedSAM \cite{medsam} & \xmark & \cmark & \xmark \\
Medical SAM Adapter \cite{medsamadapter} & \xmark & \cmark & \xmark \\
SonoSAM \cite{sonosam} & \xmark & \cmark & \xmark \\
SAMUS \cite{samus} & \xmark & \cmark & \cmark \\
SAMed \cite{samed} & \cmark & \xmark & \xmark \\
AutoSAM \cite{autosam} & \cmark & \xmark & \cmark \\
AdaptiveSAM \cite{adaptivesam} & \cmark & \cmark & \xmark \\
\textbf{S-SAM (Ours)} & \cmark & \cmark & \cmark \\
\hline
\end{tabular}
}
\label{method_comparison}
\end{wraptable}
To perform this, they tune the biases of the encoder network and keep the decoder fully trainable while also adding a text affine layer, producing good results for surgical scene segmentation. Using text prompts can be considered as requiring no expert involvement since simply the label names can serve as a valid text prompt. Hence, in S-SAM, we also perform text-prompted segmentation. However, our method does not require the mask decoder of SAM to be tuned. Hence, it is more efficient than AdaptiveSAM. Furthermore, by tuning the singular values instead of biases, our model does a better job of learning the domain shift.
In summary, an appropriate adaptation method for SAM should minimize expert involvement. It should allow for some form of prompting to facilitate interactivity as needed. Finally, it should be as efficient as possible. Our method satisfies all of these requirements, as shown in Table \ref{method_comparison}.

\section{Proposed Method}

\begin{wrapfigure}{r}{0.65\textwidth}
  \centering
  \includegraphics[width=0.65\textwidth]{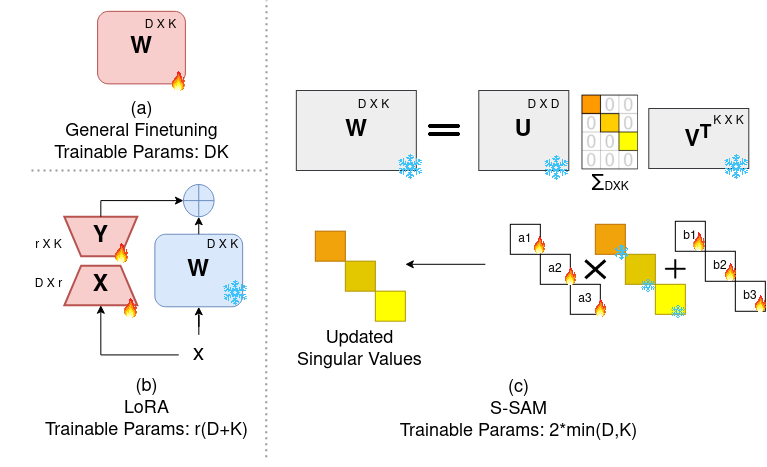}
    \caption{Comparison of different fine-tuning methods. (a) Naive fine-tuning (b) LoRA (c) Our approach only tunes the singular values and is even more efficient than LoRA.}
  \label{fig:param_count}
\end{wrapfigure}

\noindent\textbf{SVD-based Tuning: }The image encoder in SAM is a chain of \(N\) blocks, each of which has a multi-head attention module followed by a Multi Layer Perceptron (MLP), which perform the following computation
\begin{equation}
    qkv = (W_{qkv}^n)x + b_{qkv}^n,\;\; o_n = (W_{M}^n)x + b_{M}^n,
\label{qkv}
\end{equation}
where $q,k$ and $v$ denote the query, key and value associated with the multi-head attention and $M$ denotes the MLP layer. $o$ denotes the output of the block and $x$ denotes the input to a module. Here, $n$ is used to index the number of blocks (\(N\)). Note that $W\in \mathbb{R}^{D\times K}$ and $b\in \mathbb{R}^{D\times 1}$ denote the weights and biases of the respective modules. Here, \(D\) represents the dimension of the input to the module (multi-head attention or MLP) and \(K\) represents the output dimension of the respective module. These weights are primarily responsible for learning how to encode natural images after being extensively trained on them. To adapt them to the medical domain, we propose tuning W as follows:
\begin{equation}
    W \leftarrow  U ReLU(A\odot\Sigma+B) V^T, \;\;\text{where}\; W = U\Sigma V^T, 
\end{equation}
where \(W=U \Sigma V^T\) denotes the Singular Value Decomposition (SVD) of the weight matrix \(W\) and \(\odot\) represents the element-wise multiplication.  \(A\) and \(B\) are matrices with the same shape as \(\Sigma\) with non-diagonal entries as $0$. 
\begin{wrapfigure}{r}{0.5\textwidth}
  \centering
  \includegraphics[width=0.5\textwidth]{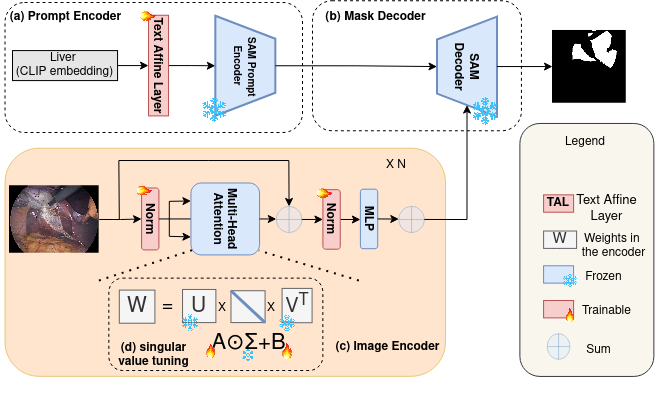}
    \caption{S-SAM Architecture. The image encoder weights are modified by performing a transformation over their singular values. Other trainable parameters include the layernorms and positional embeddings in the encoder and the Text Affine Layer (TAL). Everything else is frozen and initialized with SAM's pre-trained checkpoint.}
  \label{fig:arch}
\end{wrapfigure}
In other words, we compute the singular values of any weight matrix in the image encoder and learn a non-linear transformation over them. Since \(A\) is multiplied elementwise, it represents scaling of the singular values while \(B\) is added and thus represents shifting of the singular values. Finally, to maintain the positive semidefinite nature of the \(\Sigma\) matrix, we apply a ReLU operation. This operation is illustrated in Figure \ref{fig:param_count} (c).

\noindent\textbf{S-SAM - Architecture: }An overview of S-SAM's architecture can be found in Figure~\ref{fig:arch}. S-SAM consists of an image encoder, prompt encoder and the mask decoder. An additional module called the Text Affine Layer (TAL) is also added before the prompt encoder. The inputs to S-SAM include an image and a text prompt (which can be the label name itself) and the output is a mask over the region described by the text. The image encoder of S-SAM is SAM's original image encoder, albeit with all the weights modified as described in the previous subsection. The image is fed into this modified encoder which outputs the image embeddings. Similarly, for the text prompt, we first obtain the CLIP embedding. However, both CLIP and SAM were trained on natural images and hence might not capture the medical labels correctly. Therefore, we pass the CLIP embedding through TAL (a learnable one-layer MLP), before passing it to the prompt encoder, which outputs the prompt embeddings. Finally, we use the mask decoder to fuse the image and prompt embeddings to generate the mask of interest.

In this setup, all the weights of the image encoder, prompt encoder and mask decoder are initialized with SAM's pre-trained checkpoint. The non-zero elements of \(A\) corresponding to every weight matrix are initialized with one and \(B\) is initialized with zeros. During training, CLIP, the mask decoder and the prompt encoder are completely frozen. All the weights of the image encoder are frozen and only the transform parameters \(A\) and \(B\) are learnable. In addition, positional embeddings are trainable to allow training with smaller resolutions and layernorm layers are trainable to better adapt to the new domain. The positional embeddings in SAM's checkpoint expect the input image size to be $1024 \times 1024$. Hence, many adaptation techniques upsample the images to this scale. However, this significantly adds on to the memory requirements of these approaches. We replace SAM's positional embeddings with learnable embeddings that can facilitate training with lower image sizes. To retain the information contained in SAM, we initialize these by performing an AveragePool operation over the embedding weights present in SAM's checkpoint to bring them to the required size. Finally, the TAL is also trainable.


\noindent\textbf{Comparison with LoRA: }Low Rank Adaptation (LoRA) is a highly effective technique for fine-tuning large language models for various applications \cite{lora,lora_app1,lora_app2}. This concept involves adding a product of two low-rank matrices $X\in \mathbb{R}^{D\times r}$ and $Y\in \mathbb{R}^{r\times K}$ to W as follows:
$
    W \leftarrow W + XY.
$
However, this may lead to underfitting if the learnt subspace is smaller than required \cite{samed}. Thus, the effectiveness of this method largely depends on the rank of the approximated matrices. In S-SAM, all the singular values are tuned resulting in a full-rank computation, thus avoiding this problem. Furthermore, LoRA involves fine-tuning all the parameters of the low-rank matrices while S-SAM only tunes the singular values. This makes our approach more efficient than LoRA. Assuming $W$ is $D\times K$, the rank of \(W\) is given by $\min(D,K)$. In LoRA, the learnt matrices $X \in \mathbb{R}^{DXr}$ and $Y \in \mathbb{R}^{r \times K}$ have rank $r \ll \min(D,K)$. This makes the number of trainable parameters equal to \(r(D+K)\) as shown in Figure \ref{fig:param_count}. On the other hand, S-SAM only tunes the singular values of \(W\) by learning the scale $A$ and shift $B$ parameters. Hence, the number of trainable parameters amounts to \(2\times\min(D,K)\), which is lesser than LoRA.

\section{Experimental Results}
\noindent\textbf{Datasets: }
We evaluate S-SAM on the following five medical imaging datasets corresponding to different modalities. (i) CholecSeg8k \cite{cholecseg8k} consists of endoscopic surgery images with 12 classes of interest including surgical instruments, organs and tissues. (ii) The 
abdominal ultrasound dataset \cite{ultrasound} consists of simulated ultrasound images and has 8 classes of interest denoting different organs and bones. The testing data consists of a mix of real and synthetic images. (iii) ChestXDet \cite{chestxdet} has x-ray images with 13 classes of interest representing malignancies in the chest region. (iv) The LiTS Dataset \cite{lits_kaggle} consists of Computed Tomography (CT) images with the classes of interest being the liver and the tumor region. S-SAM takes in a 2D image as an input. Hence, we use its derived dataset having 2D slices found at \cite{lits_kaggle}. (v) The GLAS challenge dataset \cite{glas} comprises of histology images. There is only one class of interest, namely the glands, which needs to be segmented. The rest of the experimental setup is outlined in the supplementary.
\begin{wrapfigure}{r}{0.65\textwidth}
  \centering
  \includegraphics[width=0.65\textwidth]{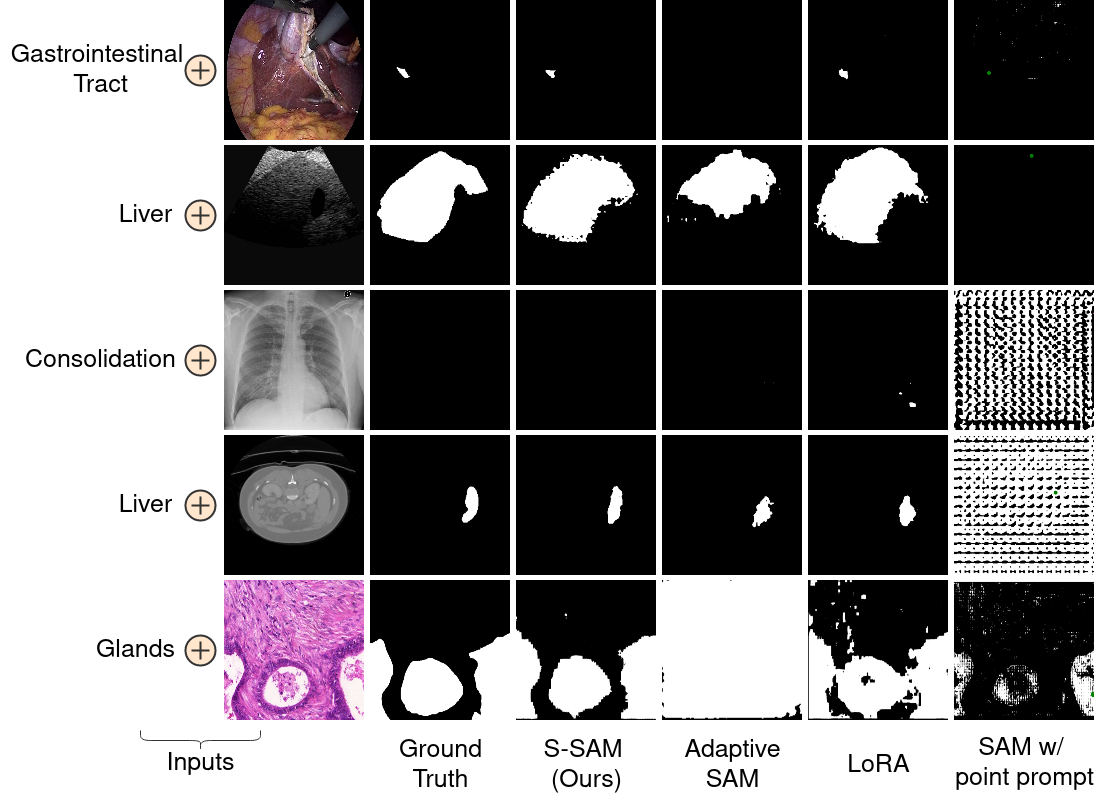}
    \caption{A qualitative comparison among different methods. From the top, the rows represent CholecSeg8k, Ultrasound, ChestXDet, LiTS, and GLAS, respectively. The green dot in the last column denotes the point prompt used to query SAM.}
  \label{fig:results}
\end{wrapfigure}

\noindent\textbf{Results: }In clinical practice, a blank mask, corresponds to the label of interest that is not present in the image. Note that a blank mask is a valid prediction. For example, if a CT scan of a normal liver is queried with the text prompt "Tumor", it should output an empty mask. In the classical definition of DICE Score (DSC), such cases have undefined score and hence, ignored. Hence, for each of the datasets, we evaluate S-SAM using the Dice Score as defined by Rahman et al. \cite{aiman} who give a DSC of 1 to cases where the predicted mask and the label are both completely blank. 


\begin{table}
\centering
\caption{Results on ChoecSeg8k. Some methods group certain labels into one category and report the groupwise results, denoted by multi-columns numbers.}

\setlength{\tabcolsep}{3pt}
\resizebox{1\columnwidth}{!}{%
\begin{tabular}{|c|c|c|c|c|c|c|c|c|c|c|c|c|c|}
\hline
\multicolumn{14}{|c|}{GB-Gall Bladder, AW-Abdominal Wall, GT-GI Tract, CD-Cystic Duct, LHEC-L Hook Electrocautery, HV-Hepatic Vein, CT-Connective Tissue, LL-Liver Ligament.}\\
\hline
Method &
\multicolumn{13}{c|}{Object wise DSC}\\
\hline
& Fat & Liver & GB & AW & GT & Grasper & LHEC & Blood & HV & CT & LL & CD & Avg. \\
\hline
\textbf{Traditional DL Methods} & & & & & & \multicolumn{2}{c|}{} & \multicolumn{5}{c|}{} & \\
U-Net\cite{unet1} & 0.87 & 0.52 & 0.40 & 0.73 & 0.26 & \multicolumn{2}{c|}{0.52} & \multicolumn{5}{c|}{0.08} & 0.48 \\



UNetR \cite{trans2} & 0.88 & 0.74 & 0.42 & 0.76 & 0.35 &\multicolumn{2}{c|}{0.71} & \multicolumn{5}{c|}{0} & 0.55 \\

TransUNet \cite{transunet} & 0.83 & 0.43 & 0.77 & 0.35 & 0.43 & 0.70 & 0.55 & 0.61 & 0.82 & 0.57 & 0.72 & 0.64 & 0.62 \\

MedT \cite{medT}& 0.81 & 0.39 & 0.56 & 0.34 & 0.25 & 0.48 & 0.71 & 1 & 0.70 & 0.69 & 0 & 0.89 & 0.57 \\
\hline
\textbf{SAM based methods}&&&&&&&&&&&&&\\
SAM w/ text prompt \cite{sam}& 0.05 & 0 & 0.02 & 0 & 0 & 0.01 & 0.04 & 0.01 & 0.14 & 0.01 & 0.14 & 0.01 & 0.04\\

SAM w/ point prompt \cite{sam} & 0.17 & 0.23 & 0.07 & 0.30 & 0.10 & 0.22 & 0.63 & 1 & 0.70 & 0.43 & 1 & 1 & 0.49 \\

SAM with full finetuning \cite{sam} & 0 & 0 & 0.02 & 0 & 0.09 & 0.13 & 0.38 & 0.91 & 0.7 & 0.38 & 1 & 0.91 & 0.38 \\

MedSAM \cite{medsam} & 0 & 0 & 0.02 & 0 & 0.08 & 0.15 & 0.46 & 1 & 0.69 & 0.38 & 1 & 1 & 0.40 \\

SAMed \cite{samed} & 0 & 0 & 0.03 & 0 & 0.13 & 0.19 & 0.46 & 1 & 0.69 & 0.38 & 1 & 1 & 0.41 \\

AdaptiveSAM \cite{adaptivesam}& 0.85 & 0.71 & 0.37 & 0.80 & 0.10 & 0.20 & 0.70 & 1 & 0.70 & 0.38 & 1 & 1 & 0.64 \\

Low Rank Adaptation of SAM \cite{lora} & 0.87 & 0.72 & 0.45 & 0.76 & 0.42 & 0.20 & 0.48 & 0.97 & 0.70 & 0.70 & 0.6 & 0.97 & 0.65 \\

S-SAM (Ours) & 0.89 & 0.71 & 0.51 & 0.73 & 0.43 & 0.30 & 0.63 & 1 & 0.71 & 0.56 & 1 & 1 & \textbf{0.71}\\
\hline
\end{tabular}}

\label{cholec}
\end{table}

\begin{figure*}[h!]
  \centering
  \includegraphics[width=\linewidth]{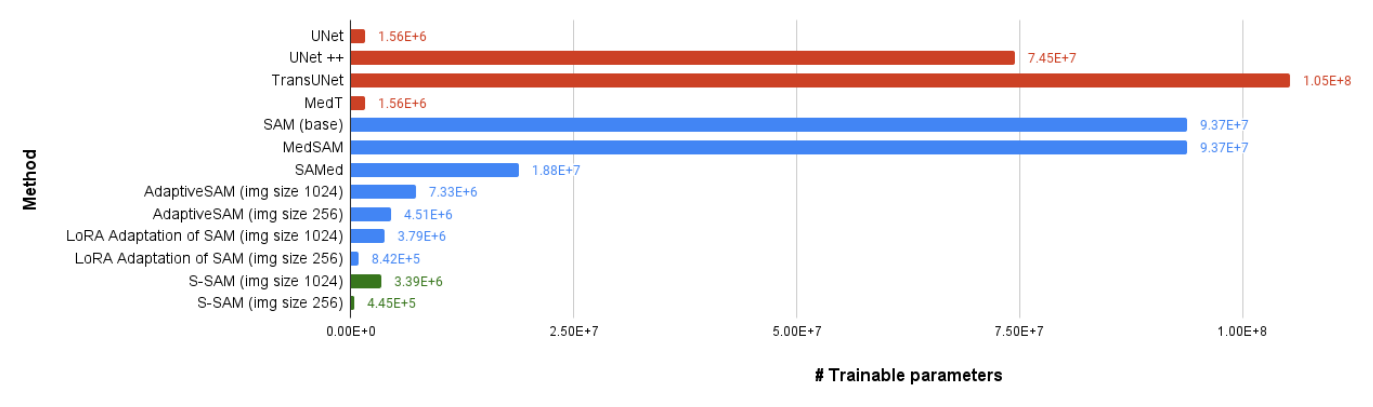}
     \caption{A comparison among different methods based on the number of parameters trained. The red bars indicate traditional DL-based segmentation methods. Blue bars indicate SAM-based methods and green bars indicate our method. The numbers to the right of each bar denote the number of trainable parameters.}
  \label{fig:num_train}
\end{figure*}

We present quantitative results for CholecSeg8k, LiTS, and GLAS in Tables \ref{cholec}, \ref{lits}, and \ref{glas} respectively. Due to space constraints, results on Abdominal Ultrasound and ChestXDet are in supplementary Tables 2 and 3 respectively. On average, we achieve a significant improvement of 6-7\% on CholecSeg8k, Ultrasound, and GLAS over the existing state-of-the-art (SOTA), and 1\% improvement on the other two datasets. However, note that our method achieves on par performance with a significantly lesser number of parameters. From the tables, we see that adapting foundation models like SAM improves performance over existing DL-based methods. Furthermore, we compare S-SAM with zero-shot SAM in the first two rows in SAM-based methods in the tables. SAM is not trained on medical images and hence performs poorly when prompted with the label name as text prompt without any finetuning. However, with a point prompt, it can still segment to some extent. Similarly, MedSAM also performs well on certain objects, but not as well as adaptation methods, showing the requirement of tuning on new datasets. We also observe that full finetuning of SAM overfits to the data, resulting in a lower test performance. Finally, we show improvements over various SAM-based adaptation methods. A comparison based on the number of parameters is provided in Figure \ref{fig:num_train}, which shows that S-SAM requires a significantly lower number of parameters than these adaptation methods, while also outperforming them on all five datasets. In comparison to SAM, there is a 99.6\% reduction in the number of trainable parameters while in comparison to AdaptiveSAM, there is 90\% reduction. Similarly, S-SAM trains 50\% lesser parameters than LoRA. All results with S-SAM have a p-value of at most $10^{-8}$ wrt other methods, which shows statistical significance.
\begin{wraptable}{r}{0.5\textwidth}
\centering
\caption{Ablation analysis on the components of S-SAM.}
\label{component_ablation}
\resizebox{0.5\textwidth}{!}{%
\begin{tabular}
{|c|c|c|c|c||c|}
\hline
Tuning Pos Embeds & Tuning Layernorm & TAL & Scaling & Shifting & Avg DSC\\
\hline

& & & & & 0.04\\
\cmark & & & & & 0.04\\
\cmark & \cmark & & & & 0.50\\

 \cmark & \cmark & \cmark & & & 0.52\\
 \cmark & \cmark & \cmark & \cmark & & 0.54 \\
\cmark & \cmark & \cmark & & \cmark & 0.64 \\
\cmark & \cmark & \cmark & \cmark & \cmark & \textbf{0.71}\\

\hline
\end{tabular}
}
\end{wraptable}
\noindent \textbf{Qualitative Results: } In Figure \ref{fig:results}, we present sample results of S-SAM and other methods. S-SAM is able to segment smaller objects like the GI Tract that is missed by Adaptive SAM or SAM as seen in row 1 of the Figure. In addition, S-SAM is also able to produce blank masks when a queried object of interest is not present, as seen in row 3. The effectiveness of adaptation methods can be visually represented through all the cases, where zero-shot SAM produces gibberish results, unlike the adaptation methods. 

\begin{table}
\parbox{.45\linewidth}{
\centering
\caption{Results on LiTS}
\label{lits}
\resizebox{0.45\columnwidth}{!}{%
\begin{tabular}
{|c|c|c|c|}
\hline
Method &
\multicolumn{3}{c|}{Objectwise DSC}\\
\hline
& Liver & Tumor & Avg. \\
\hline
\textbf{Traditional DL Methods} & && \\
UNet \cite{unet1}& 0.77 & 0.65 & 0.71\\
SegNet \cite{segnet}& 0.76 & 0.64 & 0.70\\
KiuNet \cite{kiunet}& 0.80 & 0.71 & 0.76\\
DeepLab v3+ \cite{deeplabv3p} & 0.85 & 0.68 & 0.77\\
\hline
\textbf{SAM based Methods} & & &\\
SAM w/ text prompt \cite{sam} & 0.04 & 0 & 0.02 \\
SAM w/ point prompt \cite{sam} & 0.05 & 0 & 0.03 \\
SAM with full finetuning \cite{sam} & 0.05 & 0.86 & 0.5 \\
MedSAM \cite{medsam} & 0.06 & 0.01 & 0.04 \\
SAMed \cite{samed} & 0.61 & 0.91 & 0.76\\
AdaptiveSAM \cite{adaptivesam} & 0.80 & 0.86 & 0.83\\
Low Rank Adaptation of SAM \cite{lora} & 0.82 & 0.84 & 0.83\\
S-SAM (Ours) & 0.85 & 0.83 & \textbf{0.84} \\
\hline
\end{tabular}
}
}
\hfill
\parbox{.45\linewidth}{
\centering
\caption{Results on GLAS.}
\label{glas}
\resizebox{0.41\columnwidth}{!}{%
\begin{tabular}
{|c|c|}
\hline
Method & Objectwise DSC \\
\hline
& Glands \\
\hline
\textbf{Traditional DL Methods} & \\
UNet \cite{unet1}& 0.52 \\
SegNet \cite{segnet}& 0.84 \\
clDice \cite{cldice}& 0.85 \\
PointRend \cite{pointrend} & 0.88 \\
TI-Loss \cite{tiloss}& 0.88 \\
\hline
\textbf{SAM Based Methods} & \\
SAM w/ text prompt \cite{sam}& 0.01\\
SAM w/ point prompt \cite{sam}& 0.19\\
SAM with full finetuning \cite{sam}& 0.85\\
MedSAM \cite{medsam} & 0.21\\
SAMed \cite{samed} & 0.65 \\
AdaptiveSAM \cite{adaptivesam} & 0.66\\
Low Rank Adaptation of SAM \cite{lora} & 0.83 \\
S-SAM (Ours) & \textbf{0.90} \\
\hline
\end{tabular}
}
}
\end{table}
\noindent {\bf{Component-wise Ablation:}}
S-SAM has three major modified components over SAM, namely the Text Affine Layer (TAL), scaling matrix \(A\), and the shifting matrix \(B\), as well as other modifications like training the positional embeddings and layernorms of the model. To assess the importance of each of these, we start with SAM and note the rise in performance on CholecSeg8k when each component is added. Results for this study are tabulated in Table \ref{component_ablation}. 

 The first row in the table shows SAM's zero-shot performance without modifications. Tuning only the positional embeddings of the encoder doesn't improve performance. However, tuning layernorm layers significantly boosts DSC by around 46\%, consistent with domain adaptation research where norm layers contribute to domain bias \cite{da_bn}. Adding TAL further improves performance. Shifting and scaling, added one by one, both enhance performance, with shifting appearing more important for modeling domain shift, resulting in a 12\% increase. Finally, with all components, we achieve S-SAM with the best performance.

\section{Conclusion}
In this paper, we presented S-SAM - an efficient adaptation of SAM for medical images, which is realized by tuning the singular values of the weight matrix. We show that S-SAM performs on par or outperforms existing methods on five publicly available medical image segmentation datasets with significantly lower amount of trainable parameters and allows the use of label names as prompts. We find that S-SAM could be affected by class-size disparities, as seen from its performance on specific classes. A potential future direction towards improving S-SAM could be using additional loss functions or weighted loss functions to reduce the effect of the disparities.

\section{Disclosure of Interests}
This research was supported by a grant from the National Institutes of Health, USA; R01EY033065. The content is solely the responsibility of the authors and does not necessarily represent the official views of the National Institutes of Health. The authors have no competing interests in the paper 

%
%
%
\bibliographystyle{splncs04}
\bibliography{bibliography_v2}

\end{document}


\title{Supplementary - S-SAM: SVD-based Fine-Tuning of Segment Anything Model for Medical Image Segmentation
}
%
%
%
%

%
\author{Jay N. Paranjape\inst{1,}\Letter \and
Shameema Sikder\inst{2,3} \and
S. Swaroop Vedula\inst{3} \and
Vishal M. Patel\inst{1}
}
\authorrunning{J. Paranjape et al.}
\institute{Department of Electrical and Computer Engineering, The Johns Hopkins University, Baltimore, USA \\
\email{jparanj1@jhu.edu}
\and
Wilmer Eye Institute, The Johns Hopkins University, Baltimore, USA \and Malone Center for Engineering in Healthcare, The Johns Hopkins University, Baltimore, USA
}
\maketitle            
\noindent\textbf{Preliminaries: }Segment Anything Model or SAM \cite{sam} was recently proposed as a foundational model for prompt-guided image segmentation. These prompts can be (i) foreground/background points that are vectorized using positional embeddings, (ii) bounding boxes that can be represented by the positional embeddings of their corner points, (iii) masks that can be encoded through a Convolutional Neural Network (CNN) or (iv) text that uses CLIP \cite{clip} embeddings. Note that support for the text-based prompts is not included in SAM's released codebase. Prompt-guided learning is facilitated through separate encoders for the image and the prompt, which are then fused using a mask decoder module. The image encoder in SAM is a Vision Transformer (ViT) \cite{vit} that is pre-trained using the Masked Auto-Encoder (MAE) strategy \cite{mae} and is responsible for the majority of the memory consumption. The mask decoder is a lightweight transformer-based component while the prompt encoder embeds the different prompts as described earlier and combines them.\\

\noindent {\bf{Number of Singular Values Tuned:}}
We conduct experiments to check
\begin{wraptable}{r}{0.3\textwidth}
\centering
\caption{Ablation analysis on percentage of singular values tuned.
}
\label{svdpercent}

\resizebox{0.3\textwidth}{!}{%
\begin{tabular}
{|c|c|}
\hline
Percent of singular values tuned &
Avg. DSC\\
\hline

1 & 0.49\\
10 & 0.54\\
50 & 0.56\\
100 & 0.71\\
\hline
\end{tabular}
}
\end{wraptable}
 whether all the singular values need to be tuned to model the required domain shift. As seen in Table \ref{svdpercent}, we tune only the top \(k\)\% values, where \(k \in \{1,10,50, 100\}\). However, we see a significant drop in performance on reducing \(k\) on the CholecSeg8k dataset. This is expected since the eigenvectors corresponding to the top singular values for the natural image domain might not be the most relevant vectors for the new medical domain. Hence, best performance is observed when \(k\) is 100.

\noindent\textbf{Experimental Setup: }
We use the 'ViT-base' backbone checkpoint from SAM for initializing our model, while the weights of the TAL network are initialized using the default settings of Pytorch (Kaiming Uniform). We apply augmentations including random rotation (\(\pm 10^{\circ}\)) with \(0.5\) probability, random saturation change with a scale of \(2\) with \(0.2\) probability, and random brightness change with a scale of \(2\) with \(0.5\) probability during training for all input images, followed by the normalization used in SAM. All training is done with the AdamW optimizer with a learning rate of 1e-4, on a single Nvidia RTX A6000 GPU. The memory requirement for a given training instance is less than 12 GB when the image is resized to $256 \times 256$. The loss function used for all experiments is the sum of dice loss and focal loss between the ground truth label and the predicted mask.\\

\begin{table}
\centering
\caption{Results on Abdominal Ultrasound.}
\setlength{\tabcolsep}{3pt}
\resizebox{0.8\columnwidth}{!}{%
\begin{tabular}
{|c|c|c|c|c|c|c|c|c|c|}
\hline
Method &
\multicolumn{9}{c|}{Objectwise DSC}\\
\hline
& Liver & Kidney & Pancreas & Vessels & Adrenals & \footnotesize{Gall Bladder} & Bones & Spleen & Avg. \\
\hline
\textbf{Traditional DL methods}&&&&&&&&&\\
UNet& 0.28 & 0.37 & 0.11 & 0.16 & 0.85 & 0.08 & 0.17 & 0.14 & 0.27\\

TransUNet& 0.18 & 0.09 & 0.03 & 0.03 & 0 & 0.11 & 0.05 & 0.02 & 0.08\\

MedT \cite{medT}& 0.18 & 0.03 & 0.27 & 0.10 & 0.85 & 0.15 & 0.02 & 0.08 & 0.21\\

\hline
\textbf{SAM based methods}&&&&&&&&&\\

SAM w/ text prompt& 0.17 & 0.20 & 0.72 & 0.21 & 0.44 & 0.65 & 0.67 & 0.63 & 0.46\\

SAM w/ point prompt & 0.11 & 0 & 0.01 & 0 &0.01 &0.01 & 0.01 & 0.01 & 0.02\\

SAM with full finetuning & 0.21 & 0.48 & 0.67 & 0.56 & 0.81 & 0.69 & 0.54 & 0.53 & 0.56\\

MedSAM& 0.14 & 0.03 & 0.01 & 0.01 & 0 & 0.01 & 0 & 0.02 & 0.03\\

SAMed& 0.20 & 0.50 & 0.61 & 0.56 & 0.82 & 0.63 & 0.54 & 0.54 & 0.55\\

AdaptiveSAM& 0.36 & 0.30 & 0.50 & 0.40 & 0.86 & 0.63 & 0.67 & 0.54 & 0.53 \\

Low Rank Adaptation of SAM & 0.43 & 0.35 & 0.45 & 0.61 & 0.90 & 0.59 & 0.67 & 0.67 & 0.58 \\

S-SAM (Ours) & 0.32 & 0.52 & 0.80 & 0.61 & 0.91 & 0.75 & 0.67 & 0.43 & \textbf{0.63} \\

\hline
\end{tabular}}
\label{ultrasound}
\end{table}

\begin{table}
\centering
\caption{Results on ChestXDet.}
\setlength{\tabcolsep}{3pt}
\resizebox{0.8\columnwidth}{!}{%
\begin{tabular}
{|c|c|c|c|c|c|c|c|c|c|c|c|c|c|c|}
\hline
\multicolumn{15}{|c|}{Ef - Effusion, No - Nodule, Cm - Cardiomegaly, Fb - Fibrosis, Co - Consolidation, Em - Emphysema, Ma - Mass}\\
\multicolumn{15}{|c|}{Ca - Calcification, Pt - Pleural Thickening, Pn - Pneumothorax, Fr - Fracture, At - Atelectasis, Dn - Diffuse Node}\\
\hline
Method &
\multicolumn{14}{c|}{Object wise DSC}\\
\hline

& Ef & No & Cm & Fb & Co & Em & Ma & Ca & Pt & Pn & Fr & At & Dn & Avg. \\
\hline
\textbf{Traditional DL methods}&&&&&&&&&&&&&&\\
UNet& 0.15 & 0.08 & 0.06 & 0 & 0.13 & 0.02 & 0.95 & 0 & 0.08 & 0 & 0.50 & 0.02 & 0.02 & 0.15\\
TransUNet& 0.06 & 0.87 & 0.06 & 0.59 & 0.13 & 0.01 & 0.89 & 0 & 0.74 & 0 & 0.08 & 0 & 0 & 0.26\\
MedT & 0.06 & 0.75 & 0.08 & 0.01 & 0.10 & 0.03 & 0.12 & 0 & 0.91 & 0 & 0 & 0.37 & 0.07 & 0.19\\
\hline
\textbf{SAM based methods}&&&&&&&&&&&&&&\\
SAM w/ text prompt& 0.05 & 0.13 & 0.53 & 0.36 & 0.15 & 0.28 & 0.23 & 0.10 & 0.37 & 0.07 & 0.40 & 0 & 0.26 & 0.22\\

SAM w/ point prompt& 0.04 & 0 & 0.01 & 0.01 & 0.04 & 0.01 & 0 & 0 & 0 & 0 & 0 & 0 & 0.02 & 0.01\\

SAM with full finetuning& 0.55 & 0.88 & 0.87 & 0.086 & 0.52 & 0.93 & 0.95 & 0.93 & 0.84 & 0.93 & 0.86 & 0.92 & 0.94 & \textbf{0.84}\\

MedSAM& 0.04 & 0 & 0.02 & 0.01 & 0.04 & 0.02 & 0 & 0 & 0 & 0 & 0.02 & 0 & 0.02 & 0.01\\

SAMed & 0.50 & 0.89 & 0.90 & 0.83 & 0.50 & 0.93 & 0.94 & 0.91 & 0.83 & 0.92 & 0.85 & 0.93 & 0.93 & 0.83 \\ 

AdaptiveSAM & 0.52 & 0.88 & 0.86 & 0.86 & 0.43 & 0.93 & 0.95 & 0.91 & 0.84 & 0.93 & 0.86 & 0.94 & 0.93 & 0.83\\
Low Rank Adaptation of SAM & 0.50 & 0.89 & 0.90 & 0.83 & 0.42 & 0.93 & 0.96 & 0.91 & 0.84 & 0.93 & 0.86 & 0.94 & 0.94 & 0.83\\
S-SAM (Ours) & 0.48 & 0.89 & 0.87 & 0.86 & 0.4 & 0.93 & 0.96 & 0.94 & 0.84 & 0.94 & 0.87 & 0.94 & 0.95 & \textbf{0.84}\\
\hline
\end{tabular}}
\label{xray}
\end{table}

\bibliographystyle{splncs04}
\bibliography{bibliography_v2}